\documentclass[twoside]{article}

\usepackage[accepted]{aistats2023}

\usepackage[table,xcdraw]{xcolor}
\usepackage{graphicx}
\usepackage{bbm}
\usepackage{amsmath}
\usepackage{amsfonts}
\usepackage{dsfont}
\usepackage{graphicx}
\usepackage{floatrow}
\usepackage{floatrow}
\usepackage{mathtools}
\usepackage{booktabs}
\usepackage{multirow}

\DeclareCaptionFont{Small}{\fontsize{12}{14}\selectfont}

\DeclareMathOperator{\Uniform}{Uniform}
\DeclareMathOperator{\Bernoulli}{Bernoulli}

\DeclareMathOperator{\GNN}{GNN}
\DeclareMathOperator{\Sigmoid}{Sigmoid}
\DeclareMathOperator{\ReLU}{ReLU}

\newcommand{\add}[1] {{\color{black}{#1}}} 

\begin{document}

%

%

\twocolumn[

\aistatstitle{Implicit Graphon Neural Representation}

\aistatsauthor{Xinyue Xia \And Gal Mishne \And  Yusu Wang}
\aistatsaddress{Halıcıo\u{g}lu Data Science Institute \\ University of California San Diego } ]

\begin{abstract}

Graphons are general and powerful models for generating graphs of varying size. In this paper, we propose to directly model graphons using neural networks, obtaining Implicit Graphon Neural Representation (IGNR). Existing work in modeling and reconstructing graphons often approximates a target graphon by a fixed resolution piece-wise constant representation. Our IGNR has the benefit that it can represent graphons up to arbitrary resolutions, and enables natural and efficient generation of arbitrary sized graphs with desired structure once the model is learned. Furthermore, we allow the input graph data to be unaligned and have different sizes by leveraging the Gromov-Wasserstein distance. We first demonstrate the effectiveness of our model by showing its superior performance on a graphon learning task. We then propose an extension of IGNR that can be incorporated into an auto-encoder framework, and demonstrate its good performance under a more general setting of graphon learning. We also show that our model is suitable for graph representation learning and graph generation.

\end{abstract}







\section{INTRODUCTION}






Graphs are ubiquitous in real life, from physical and chemical interactions to brain and social networks. Learning on graph data and developing statistical models for graphs have been of long-standing interest \cite{goldenberg2010survey,newman2018networks}. Classical statistical models for graph data mostly belong to parametric families, such as the stochastic block model \cite{nowicki2001estimation}, the exponential random graph model \cite{hunter2006inference}, the mixed membership model \cite{airoldi2008mixed} and many others \cite{erdHos1960evolution,hoff2002latent,soufiani2012graphlet}. However, because these models are designed to characterize a very particular type of graphs, their modeling capacity becomes limited when dealing with more complex and diverse graphs. 

The \textit{graphon} \cite{lovasz2012large,diaconis2007graph} has emerged as a non-parametric, statistical model for graphs, and its modeling capacity greatly exceeds the aforementioned parametric graph models. Graphons have recently garnered increasing interest in statistical and machine learning \cite{eldridge2016graphons,ruiz2020graphon,ruiz2021graphon} with their two-fold theoretical interpretations. A graphon can either be interpreted as the limit object of a convergent sequence of graphs, or, pertinent to our present work, as a very general model for generating unweighted graphs. To be formally introduced in Section 2, any graphon can be represented as a symmetric, measurable 2D function $W: [0,1]^2 \to [0,1]$. Adopting the function representation of graphons, one interesting question that arises is how to model graphons and learn graphons from graph data.

Prior work has developed methods for learning graphons from data, e.g. matrix completion \cite{keshavan2010matrix}, stochastic block approximation \cite{airoldi2013stochastic}, sorting-and-smoothing \cite{chan2014consistent}, universal singular value thresholding \cite{chatterjee2015matrix}, and Gromov-Wasserstein barycenters \cite{xu2019gromov}.
However these methods suffer from two common limitations. The first limitation is that these methods model the target graphon as a ``discrete" 2D step function of fixed resolution (i.e. a matrix), which substantially reduces the variety of graphon functions that can be nicely characterized. The second limitation is that these methods are designed to learn a single graphon from observed graphs. Real networks, however, may vary with time, space, or other general higher dimensional latent parameters. The existing methods are not applicable to the more complex and dynamic settings of graphon learning.

\paragraph{Present work.}
In this paper, we address the above limitations and present \textit{Implicit Graphon Neural Representation} (IGNR), a graphon learning framework that can represent graphons up to arbitrary resolutions, and is applicable to the more general settings of graphon learning. IGNR utilize a powerful and flexible tool,  implicit neural representation \cite{sitzmann2020implicit}, to construct a novel model for graphons. Implicit neural representation refers to using neural networks to represent implicitly defined signals. It arises as an attractive paradigm because of its potential to model various complex signals, and its ability to re-sample the signal at arbitrary resolutions. Drawing inspiration from such a paradigm, we observe that we can treat graph data as signals sampled at different resolutions (giving rise to graphs of different sizes) from some graphons. In other words, we can consider graphons as functions defined implicitly through graph data, and thus model graphons with implicit neural representation. Such a neural representation for graphons resolves the first limitation mentioned previously, because a neural network can characterize a much broader family of functions than that of a fixed-resolution step function. The second limitation of learning graphons in more general settings can also be readily solved because given the graphon modeled as a neural network, we can easily incorporate a latent space into the model through an auto-encoder framework, and utilize the latent space to capture latent parameters that govern the changing graphons. To summarize, our contribution is two-fold:
\begin{itemize}
\item We propose IGNR, a novel framework for modeling and learning graphons that enables resolution-free representation of graphons. Extending IGNR, we further propose conditional-IGNR \add{(c-IGNR)}, which can be incorporated into an auto-encoder to solve more general graphon learning tasks.

\item We validate the effectiveness of our framework by showing that it excels in both classical and more general settings of graphon learning.  We also show that our framework produces meaningful graph embeddings and offers a convenient model for generating arbitrary sized graphs.
\end{itemize}

\subsection*{Related Work} 

\textit{Graphon Learning}. Classical graphon learning methods \cite{airoldi2013stochastic,chan2014consistent,keshavan2010matrix,chatterjee2015matrix} 
often make an important assumption
on the input graphs---these graphs need to be ``well-aligned", which means the correspondence between nodes should be given so that all input graphs can first be arranged into some common node ordering. If the node arrangements are not given, heuristics will be deployed, for example, by sorting the nodes with empirical degrees \cite{chan2014consistent}. Such a pre-processing step may induce unwanted error due to undesired matching, for example, when the graph nodes cannot be arranged in some common ordering. The work of \cite{xu2021learning} bypasses the problem of node arrangements by employing the Gromov-Wasserstein (GW) distance, which is permutation invariant to the node orderings of graphs. They use the GW barycenter of the input graphs as the estimate of the target graphon, but they still rely on a piece-wise constant approximation for the target graphon. The resolution of the barycenter is a parameter that needs to be manually set. Our work also leverages the GW distance by incorporating it as a reconstruction loss to train IGNR. 
\add{Also relevant to our work is GNAE \cite{xu2021graphon}, which uses a linear factorization model to represent a graphon, with the linear coefficients obtained from a graph neural network encoder. 
The graphon factors of GNAE are piece-wise-constant functions (matrices) at fixed resolutions; in contrast, IGNR learns a single, resolution-free representation of the graphon. 
}

\textit{Gromov-Wasserstein (GW) Distance.}
Optimal Transport (OT) has been used by the machine learning community under various scenarios, including unsupervised learning \cite{arjovsky2017wasserstein,schmitz2018wasserstein}, classification \cite{frogner2015learning}, natural language processing \cite{kusner2015word}, and many others. 
\add{Traditional OT loss, however, suffers from the limitation that it is not invariant to important families of invariance, such as re-scaling, translation, or rotations \cite{peyre2019computational}, and thus is not directly applicable to shape matching or comparing structured data like graphs. The GW distance proposed by \cite{memoli2011gromov} extends the original OT formulation and enables comparing distributions defined on different spaces without requiring the definition of a family of invariances.}
In particular, the comparison of the structural information between the different spaces is encoded in the OT problem. The GW distance has thus been leveraged in a sequence of following works as a distance between graphs with applications to computing graph barycenters \cite{peyre2016gromov,xu2021learning}, graph node embedding \cite{xu2019gromov}, graph partitioning \cite{xu2019scalable}, linear and non-linear graph dictionary learning \cite{vincent2021online,xu2020gromov}, and supervised graph prediction \cite{brogat2022learning}. The works of \cite{vincent2021online,xu2020gromov} are most relevant to us in the sense that they also utilize the GW distance as a reconstruction loss. However, they differ from our approach as their goal is to learn factorization models, whereas our goal is to learn a neural graphon representation.
We note that in general, computing the GW distance is hard (in the discrete setting it is often modeled as a quadratic assignment problem). 
However, we will leverage recent advances in computational optimal transport and use fast iterative algorithms such as the conditional gradient algorithm \cite{titouan2019optimal}, and proximal gradient algorithm \cite{xu2019gromov} to efficiently compute a good solution.

\add{
\textit{Deep Generative Models for graphs.} 
Variational autoencoders (VAE) \cite{kingma2013auto}, generative adversarial networks (GAN) \cite{goodfellow2014gan}, and deep autoregressive models have been applied in the past for graph generation tasks.
VAE node-based approaches (and more generally latent position models \cite{smith2019geometry}) such as VGAE \cite{kipf2016variational} learn latent node embeddings to generate edge probabilities. This approach is limited to learning from a single fixed-sized graph and hence can only generate new graphs of the same size. 
Graph-level VAE (GraphVAE \cite{simonovsky2018graphvae}) incorporates pooling to instead generate a graph-level embedding, and maps the graph embedding directly to an edge probability matrix by a decoder. 
However, it also cannot naturally deal with varying graph sizes and has to specify a maximum graph size for the decoder and use masking. This issue is common for most follow-up work in this direction. Efficient graph matching without knowing node correspondence (which is needed in reconstruction loss) is also a challenge.
In comparison, our (c-)IGNR when applied for graph generation, does not use node-level embeddings; it naturally solves the graph size generalizability issue through learning a graphon neural representation; and uses the Gromov-Wasserstein loss as an alternative to mitigate the graph matching. 
The GAN approach to graph generation avoids graph matching by using a permutation invariant discriminator, but the difficulty in minimax optimization remains its major limitation. Autoregressive approaches such as GraphRNN [You et al., 2018] and GRAN [Liao et al., 2019] solve the graph size issue by learning complex recurrent models to iteratively generate new nodes and edges. Permutation invariance remains a challenge for such approaches, and they choose to specify certain orderings over the graph nodes for training. In contrast, our IGNR enjoys the merit of conceptual and modeling simplicity (using a simple MLP to learn a graphon vs. hierarchical RNNs to learn graph generation steps).
}


\section{BACKGROUND}
\textbf{Graphon.} A graphon is a bounded, symmetric, and Lebesgue measurable function, denoted as $W: \Omega^2 \to [0,1]$, where $\Omega$ is a probability space, equipped with probability measure $\mu$. We follow the convention and set $\Omega=[0,1]$, and $\mu$ as the uniform distribution on $\Omega$. Intuitively, we can consider points on the unit line $v_i,v_j\in [0,1]$ as nodes, and $W(v_i,v_j)$ as the edge weight connecting $v_i$ and $v_j$. Given a graphon $W$, we can generate unweighted graphs of arbitrary sizes either in a stochastic or deterministic fashion. In the stochastic setting, we generate from $W$ a graph of size $N$ (represented by the adjacency matrix $A = [a_{ij}] \in \{0,1\}^{N\times N}$) by following the sampling process:
\begin{equation}
    \begin{split}
        v_i & \sim \Uniform([0,1]), \forall i = 1,..., N, \\
        a_{ij} & \sim \Bernoulli(W(v_i,v_j)), \forall i,j=1,...,N.
    \end{split}
\end{equation}
For the deterministic setting, we simply replace the random sampling of nodes $v_i$ with fixed grid (i.e. $v_i=\frac{i-1}{N}$). The graphon thus serves as a general and useful model to characterize and generate varying sized graphs. 
Moreover, the distribution on graphs defined by a graphon $W$ is unchanged by relabeling of $W$'s nodes. 

\textbf{Implicit Neural Representation.}
Implicit neural representations (INRs) have gained recent attention in the computer vision community for their conceptually simple formulation and powerful ability to represent complex signals such as images, shapes, and videos \cite{chen2019learning,sitzmann2020implicit,genova2020local,park2019deepsdf,groueix2018papier,mildenhall2021nerf}. In the common INR setting, the observed data $\mathbf{o}_i\in \mathbb{R}^p$ are considered as discrete realisations $\mathbf{o}_i=f(\mathbf{x}_i)$  of some unknown signal $f: \mathcal{X} \to \mathcal{O}$ (with $\mathcal{X}\subseteq \mathbb{R}^d$ and $\mathcal{O}\subseteq \mathbb{R}^p$) sampled at coordinates $\mathbf{x}_i \in \mathcal{X}$ for $i=1,...,n$. In the example of an image, $\mathbf{o}_i$ represents the 3-dimensional RGB value at a single pixel coordinate $\mathbf{x}_i=(x_i,y_i)$. As such, the unknown function $f$ is implicitly defined through $\mathbf{x}_i$'s and $\mathbf{o}_i$'s. 
In this case INR 
can be a neural network $f_\theta: \mathcal{X} \to \mathcal{O}$, typically a multilayer perceptron (MLP) parameterized by $\theta$, trained on the pairs $(\mathbf{x}_i,f(\mathbf{x}_i))$ to approximate $f$. 
Because $f_{\theta}$ is trained on the full continuous domain of $\mathcal{X}$, it is resolution free --- at inference time, we can evaluate $f_{\theta}$ at arbitrary points in $\mathcal{X}$ to approximate the signal value.

\textbf{Gromov-Wasserstein Distance for graphs.}
Based on the concepts of optimal transport, \cite{memoli2011gromov} proposed the Gromov-Wasserstein (GW) distance for object matching. The GW distance operates on metric measure spaces and has the important property that it is invariant to isometries of the spaces that it compares. Therefore when applying the GW distance to graphs, it is permutation invariant to the node orderings of the graphs. Formally, consider two graphs $G_1=(A_1,h_1)$ with $N_1$ nodes and $G_2=(A_2,h_2)$ with $N_2$ nodes, where $A_i \in \mathbb{R}^{N_i\times N_i}$ is the (potentially weighted) adjacency matrix, and $h_i\in\Sigma_{N_i}:=\{h\in\mathbb{R}_{N_i}^{+}|\sum_{j}h[j]=1\}$ is a histogram (i.e. probability distribution) on the $N_i$ nodes signifying their relative importance (without prior knowledge, the uniform weight $h_i=\frac{1}{N_i} \mathds{1}_{N_i}$ is taken). The 2-order squared GW distance $GW_2(G_1,G_2)$ between $G_1$ and $G_2$ is defined as:
\begin{equation}
    \begin{split}
\min_{T\in\mathcal{C}(h_1,h_2)} & \sum_{i,k=1}^{N_1}\sum_{j,l=1}^{N_2}(A_1[i,k]-A_2[j,l])^2T[i,j]T[k,l]
    \end{split}
\end{equation}
where $\mathcal{C}(h_1,h_2):=\{T\in\mathbb{R}^{N_1\times N_2}, T\mathds{1}_{N_2}=h_1, T^{\top} \mathds{1}_{N_1}=h_2 \}$ is the set of couplings between the two histograms. In the following, we assume the weights $h_1$ and $h_2$ are uniform in $GW_2(G_1,G_2)$. Note that in addition to the nice property of being permutation invariant, $GW_2$ is well defined  to compare graphs with different number of nodes, and thus it is a natural and suitable loss for graph matching. 


\section{Implicit Graphon Neural Representation (IGNR) and conditional-IGNR for graphon learning}




In this paper, we propose to leverage implicit neural representations to learn and model graphons, which we will refer to as Implicit Graphon Neural Representation (IGNR). 
We first introduce the motivation for such a representation. 
In Section 3.2, we describe the detail of IGNR for learning a single graphon from graph data. 
Next in Section 3.3, we further propose a \emph{conditional-IGNR} (c-IGNR) model to learn a parameterized family of graphons. We show how this allows us to develop an auto-encoder framework that can be potentially used for graph generative modeling and representation learning. 
Finally, we discuss how to train an IGNR with the GW loss.

\subsection{Motivation}
One of our main motivations to model graphons with INRs is to obtain resolution-free representations for graphons. 
All existing methods for graphon learning boil down to using a two-dimensional step function (a matrix with fixed $K\times K$ resolution) to represent the graphon. Such step function representations of graphons are based on the weak regularity lemma of graphon \cite{frieze1999quick}. To state the lemma, we first introduce the cut norm. Denote the space of all bounded symmetric measurable functions $\widetilde{W}: [0,1]^2 \to \mathbb{R}$ by $\widetilde{\mathcal{W}}$, and the space of graphon by $\mathcal{W}$. For $\widetilde{W}\in \widetilde{\mathcal{W}}$, the cut norm is defined as $||\widetilde{W}||_{\square} = \sup_{S,T\subseteq[0,1]}|\int_{S\times T}\widetilde{W}(x,y)dxdy|$, where the supremum is taken over all measurable subsets $S$ and $T$ of $[0,1]$. Let $\mathcal{P}=(\mathcal{P}_1,...,\mathcal{P}_K)$ be a partition of $[0,1]$ into $|\mathcal{P}|=K$ measurable subsets. We denote a step function $W_\mathcal{P}: [0,1]^2\to[0,1]$ as $W_{\mathcal{P}}(x,y)=\sum_{k,k'=1}^Kw_{kk'}\mathds{1}_{\mathcal{P}_k\times \mathcal{P}_{k'}}(x,y)$, where $w_{kk'}\in[0,1]$, and $\mathds{1}_{\mathcal{P}_k\times \mathcal{P}_{k'}}$ is the indicator function that equals 1 if $(x,y) \in \mathcal{P}_k\times \mathcal{P}_{k'}$ and equals 0 otherwise. The weak regularity lemma for graphons states that every graphon can be approximated in the cut norm by a step function up to the resolution of the step function.

\newtheorem{theorem}{Theorem}
\begin{theorem} (Weak Regularity Lemma of graphon (Lov\'asz, 2012)). For every graphon $W \in\mathcal{W}$ and $K\geq 1$, there is a step function $W_{\mathcal{P}}$ with resolution $|\mathcal{P}|=K$ such that
\begin{equation}
||W-W_P||_{\square}\leq \frac{2}{\sqrt{\log K}}    
\end{equation}
\end{theorem}

From the theorem, we can see that when using step functions to approximate graphons, the error in cut norm is inversely related to the resolution $K$, which in practice can be a small ($\sim$15) user defined value, as in \cite{xu2021learning}. Our motivation is that we want to avoid the resolution problem by using a neural network (INR), which is resolution-free, to model the graphon, instead of using matrices to model the step function approximation of the graphon. Thanks to the universal approximation theory \cite{hornik1989multilayer,hornik1991approximation}, IGNR can approximate 
any 
graphon function with arbitrary precision, 
but without dependence on resolution parameters of any sort. To the best of our knowledge, we are the first to directly use neural networks to model graphons.

\begin{figure}[h]
\includegraphics[scale=0.52]{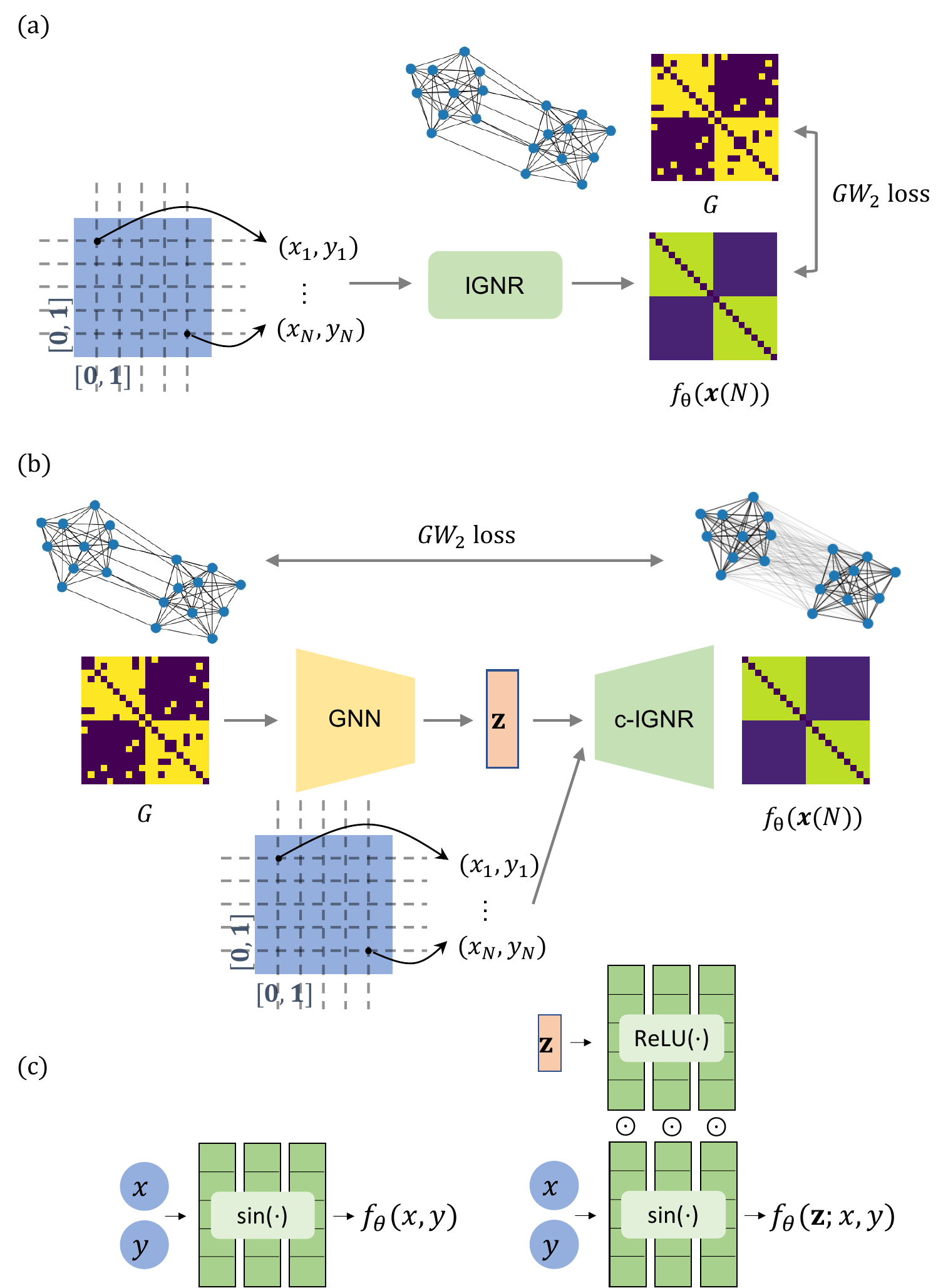}
\caption{An illustration of the (a) IGNR and (b) c-IGNR within an AE framework for graphon learning (c) left: inner architecture of IGNR, right: inner architecture of c-IGNR.}
\end{figure}

\subsection{IGNR: learning single graphons}
For the single graphon learning problem, we consider an unknown graphon $W \in \mathcal{W}$. Let $\{G_i\}_{i=1}^M$ be a set of graphs generated from $W$ using the stochastic setting. We propose to learn $W$ based on $\{G_i\}_{i=1}^M$ using a neural network parameterized by $\theta$:
\begin{equation}
    f_{\theta}: \mathbb{R}^2 \to [0,1],
\end{equation}
where $f_{\theta}$ defines a continuous function from $[0,1]^2$ to its output domain encoding edge probability. Note that although $f_{\theta}$ is defined on all of $\mathbb{R}^2$, our domain of interest for it is on $[0,1]^2$. We model $f_{\theta}$ using SIREN \cite{sitzmann2020implicit}, which is a MLP using sine activation  and has shown good performance on image data. 
Concretely, $f_{\theta}$ is a composition of $L$ hidden layers followed by an output layer with sigmoid activation:
\begin{align}
    \mathbf{h}_i &= \sin(\mathbf{W}_i\mathbf{h}_{i-1}+\mathbf{b}_i) \text{,  for  } i=1,...,L \\
    h_{L+1} &= \Sigmoid(\mathbf{W}_{L+1} \mathbf{h}_L + \mathbf{b}_{L+1})
\end{align}
with learnable weights $\theta = \{\mathbf{W}_i \in \mathbb{R}^{l_i \times l_{i-1}}, \mathbf{b}_i\in \mathbb{R}^{l_i}\}$. We set $\mathbf{h}_0=\mathbf{x} \in [0,1]^2$ as the input coordinates in $[0,1]^2$.

Given an input graph $G_i$ with $N_i$ number of nodes, we can optimize $\theta$ by minimizing the reconstruction loss $\mathcal{L}(f_{\theta}(\mathbf{x}(N_i)),G_i)$, which is chosen as the squared GW distance ($GW_2$). By abuse of notation, here we treat $\mathbf{x}$ as a coordinate sampling function that samples a grid of $N_i$ by $N_i$ coordinates in $[0,1]^2$ given $G_i$. In other words, $f_{\theta}(\mathbf{x}(N_i))$ is a $N_i$ by $N_i$ weighted adjacency matrix, whose elements represent edge connection probabilities, and we want it to be close to the input graph. As a result, the squared GW distance $GW_2(f_{\theta}(\mathbf{x}(N_i)),G_i)$ can be evaluated between the edge probability matrix (learned and sampled by IGNR) and the input graph $G_i$ (represented by its adjacency matrix $A_i$ for computation). The final graphon learning problem for IGNR given a dataset of $M$ graphs can be expressed as:
\begin{equation}
    \min_{\theta} \frac{1}{M}\sum_{i=1}^M GW_2(f_{\theta}(\mathbf{x}(N_i)),G_i)
\end{equation}
 An illustration of the graphon learning framework is shown in \add{Figure 1(a), (c)}. Note that for simplicity, we select the coordinate sampling function such that $\mathbf{x}(N_i)$ samples a set of $N_i \times N_i$ \emph{regular}-spaced coordinates in $[0,1]^2$, namely, $\mathbf{x}(N_i)=\{\mathbf{x}_{p,q}=(x_p,y_q)|x_p=\frac{p-1}{N_i},y_q=\frac{q-1}{N_i}, p=1,...,N_i,q=1,...,N_i\}$. We leave the selection of more complex sampling functions $\mathbf{x}(\cdot)$ for future work. 

\subsection{Conditional-IGNR (c-IGNR): learning parameterized family of graphons}

Instead of learning a single unknown graphon, we now consider the more general setting of learning a family of unknown graphons $\{W_{\alpha}\}_{\alpha} \subset \mathcal{W}$ that are parameterized by some unknown parameter $\alpha$. For example, when $\alpha \in [0,\infty)$, we can think of it as a time parameter, and $W_{\alpha}$ represents a generic dynamic graphon model. Let $\{G_i\}_{i=1}^M$ be a set of graphs generated from $\{W_{\alpha}\}_{\alpha}$ for different values of $\alpha$. We hope to learn for each input graph its latent representation and IGNR such that the latent space captures information of $\alpha$, and the trained IGNR can generate graphs of different sizes with similar structure to the input graph. To this end, we propose to learn a c-IGNR (\add{Figure 1(b), (c)}), which defines a conditional mapping that outputs different IGNRs ($\mathbb{R}^2\to [0,1]$) conditioned on different latent vectors ($z \in \mathbb{R}^d$). Formally, c-IGNR defines the continuous mapping:
\begin{equation}
    f_{\theta}: \mathbb{R}^d \times \mathbb{R}^2 \to [0,1]
\end{equation}
Again, $f_{\theta}$ is a neural network parameterized by $\theta$, and $d$ is the dimension of the latent space. Given the latent code $\mathbf{z}_i\in \mathbb{R}^d$ for $G_i$, $f_{\theta}(\mathbf{z}_i,\cdot)$ approximates a graphon that can generate $G_i$. There are several options to model c-IGNR, and we adopt the dual network architecture proposed by \cite{mehta2021modulated}, which was shown to be more suitable than conditioning-by-concatenation \cite{park2019deepsdf} for sine activation, and more efficient than conditional hypernetworks \cite{ha2016hypernetworks,sitzmann2020implicit}.

Concretely, c-IGNR is composed of a synthesis network and a modulation network. The synthesis network is like the ordinary IGNR, but with an additional modulation variable acting element-wise on each layer's sine activation output. It is a composition of $L$ hidden layers followed by an output layer with sigmoid activation:
\begin{align}
    \mathbf{h}_i &= \mathbf{a}_i\odot\sin(\mathbf{W}_i\mathbf{h}_{i-1}+\mathbf{b}_i) \text{,  for  } i=1,...,L \\
    h_{L+1} &= \Sigmoid(\mathbf{W}_{L+1} \mathbf{h}_L + \mathbf{b}_{L+1})
\end{align}
We set $\mathbf{h}_0=\mathbf{x} \in [0,1]^2$ as the input coordinates in $[0,1]^2$ as before. $\mathbf{W}_i \in \mathbb{R}^{l_i \times l_{i-1}}$ and $\mathbf{b}_i\in \mathbb{R}^{l_i}$ are learnable weights, and $\odot$ represents element-wise product. $\mathbf{a}_i\in \mathbf{R}^{l_i}$ is the modulation variable, exerting the influence of the latent code on IGNR.

The modulation variable $\mathbf{a}_i$ is the output of from the modulation network, which is a second MLP using ReLU activation. The modulation network takes the input graph's latent code $\mathbf{z}$ and generates the modulation variable $\mathbf{a}_i$ (for $i=1,...,L$) at each layer for the synthesis network:
\begin{align}
    \mathbf{a}_1 &:=\mathbf{h}'_{1} = \ReLU(\mathbf{W}'_{0} \mathbf{z} + \mathbf{b}'_{0})\\
    \mathbf{a}_i &:=\mathbf{h}'_i = \ReLU(\mathbf{W}'_i[\mathbf{h}_{i-1} \; \mathbf{z}]^T+\mathbf{b}'_i),
    \text{for  } i=2,...,L 
\end{align}
where $\mathbf{W}'_0\in \mathbb{R}^{l1\times d}$, $\mathbf{b}'_i\in \mathbb{R}^{l_i}$(for $i=1,...,L$), and $\mathbf{W}'_i\in \mathbb{R}^{l_i\times (d+l_{i-1})}$ (for $i=2,...,L$) are learnable weights.

Now, given the latent code $\mathbf{z}_i$ of an input graph $G_i$ with size $N_i$, and any integer $K>0$, $f_{\theta}(\mathbf{z}_i,\mathbf{x}(K))$ is the $K \times K$ weighted adjacency matrix whose elements represent edge connection probabilities, and can be intuitively considered as $G_i$'s reconstructed graph of size $K$ from its latent code $\mathbf{z}_i$. Here we treat $\mathbf{x}(\cdot)$ as the coordinate sampling function defined previously. To learn $f_{\theta}(\cdot,\cdot)$, we can treat it as a graphon decoder and incorporate it into an autoencoder (AE) framework. $f_{\theta}(\cdot,\cdot)$ constitutes a decoder because conditioning on the latent code $\mathbf{z}_i$, it outputs a graphon $f_{\theta}(\mathbf{z}_i,\cdot)$ (represented as IGNR), from which we can sample reconstructed graph of any size $K$ using coordinates $\mathbf{x}(K)$.

We complete the AE framework by adding an encoder module. We propose to use a graph neural network (GNN) parameterized by $\phi$ as an encoder ($\GNN_{\phi}$), which maps each input graph to its graph level embedding, and we treat the embedding as the latent code $z$ of the graph. Based on $M$ training graphs, the learning problem for c-IGNR in an AE framework can be expressed as:
\begin{equation}
    \min_{\phi,\theta} \frac{1}{M}\sum_{i=1}^M GW_2(f_{\theta}(\GNN_{\phi}(G_i),\mathbf{x}(N_i)),G_i)
\end{equation}

It is worth noticing that we adopt the AE framework to train c-IGNR and learn latent representations for graph data. However, once the training is done, we can separate c-IGNR from the AE and use it independently as a graph generation model that can easily generate varying sized graphs with structures similar to that of the training graph data.

\subsection{Training IGNR with GW loss}
We optimize the model parameters ($\theta$,$\phi$ in (7) and (13)) with respect to the $GW_2$ reconstruction loss by alternately updating the optimal transport coupling matrix ($T_i$) and the model parameters. Note that given fixed optimal transport matrix $T_i$, the gradient w.r.t the sampled reconstructed graph ($f_{\theta}(\GNN_{\phi}(G_i),\mathbf{x}(N_i))$, or $f_{\theta}(\mathbf{x}(N_i))$) can be easily computed, and hence we can apply back-propagation to update the network parameters ($\theta,\phi$). Given fixed network parameters, the problem of solving for the optimal transport matrix is a non-convex quadratic program, which is hard to compute. However, we can leverage recent advances in computational optimal transport and use fast iterative algorithms such as the conditional gradient algorithm \cite{titouan2019optimal}, and proximal gradient algorithm \cite{xu2019gromov} to efficiently compute the approximate solution.

\section{EXPERIMENTS}
In Section \ref{subsec:IGNRsinglegraphon}, we first demonstrate the effectiveness of IGNR 
on the task of learning a single graphon.
Motivated by the superior performance of IGNR
for learning individual graphons, we introduce the more challenging datasets for c-IGNR to learn a parameterized family of graphons. 
In Section \ref{subsec:graphonAE},
we demonstrate the strength of IGNR by showing that c-IGNR as a decoder achieves better performance than a piecewise constant baseline and state-of-the-art GW method. Finally, in Section \ref{subsec:classification} we show that the embeddings learned by c-IGNR achieves competitive graph classification result on real datasets even though the model was not designed for this purpose.

\subsection{IGNR on learning individual graphons}
\label{subsec:IGNRsinglegraphon}

\paragraph{Experiments setup.} 
We select a set of 13 different graphons (indexed by 0-12), which are considered in \cite{chan2014consistent,xu2021learning}. The definitions of these graphons are given in the supplementary material. Graphons 0-8 generate graphs that are easy to align by sorting their node degrees in strictly increasing order. Graphs generated by graphons 9-12 admit no single way to align as their degrees can be constant or non-monotonic. We test two versions of our IGNR, \textbf{IGNR-cg} and \textbf{IGNR-pg}, which solve the optimal transport matrix using the conditional gradient algorithm and the proximal gradient algorithm respectively. We compare our method with the sorting-and-smoothing (\textbf{SAS}) \cite{chan2014consistent} and the Gromov-Wasserstein Barycenter (\textbf{GWB}) \cite{xu2021learning} methods. Comparisons to other classical graphon learning methods can be found in the supplement.

We set the resolution of the ground truth graphons to be $1000 \times 1000$. For each graphon, we generate 10 graphs using the stochastic setting. To increase the difficulty of learning, these 10 graphs are of different sizes in $\{50, 77, 105, 133, 161, 188, 216, 244, 272, 300\}$. For evaluation, we use the 2-order GW distance between the ground-truth graphon and the estimated graphon as an estimation of error \add{(mean squared error could be used to estimate errors for easy-to-align graphons, see supplement)}. Because our IGNR can approximate graphons up to arbitrary resolution, at evaluation time, we sample it at resolution $1000$ as the estimated graphon. The other baselines, however, only output piecewise constant graphon estimation at a fixed resolution $K<1000$. To evaluate, we follow the procedure in \cite{xu2021learning}, which up-samples the graphon estimation to resolution $1000$ via linear interpolation as the final graphon estimation. For each graphon, we repeat the experiment in 10 trials (in each trial, we generate 10 graphs from the graphon using a different random seed and estimate the graphon by different methods), and report the mean and standard deviation of the estimation errors in Table 1.
\setlength\extrarowheight{2pt}
\begin{table}[h]
\caption{Errors of learning single graphons (each index corresponds to a different graphon)} 
\centering \small
\resizebox{\columnwidth}{!}{%
\begin{tabular}{lllll} 
\toprule
\addlinespace[8pt]
 &
  \textbf{SAS} &
  \cellcolor[HTML]{FFFFFF}\textbf{GWB} &
  \cellcolor[HTML]{FFFFFF}\textbf{IGNR-cg} &
  \cellcolor[HTML]{FFFFFF}\textbf{IGNR-pg} \\ 
\midrule
0  & 0.173$\pm$0.005 & 0.049$\pm$0.008 & 0.034$\pm$0.005 & \textbf{0.026}$\pm$0.003 \\ 
1  & 0.197$\pm$0.002 & 0.047$\pm$0.001 & 0.035$\pm$0.002 & \textbf{0.020}$\pm$0.002  \\ 
2  & 0.303$\pm$0.003 & 0.044$\pm$0.004 & 0.028$\pm$0.003 & \textbf{0.023}$\pm$0.003 \\ 
3  & 0.305$\pm$0.003 & 0.042$\pm$0.004 & 0.028$\pm$0.004 & \textbf{0.023}$\pm$0.004 \\ 
4  & 0.468$\pm$0.002 & 0.029$\pm$0.002 & 0.023$\pm$0.003 & \textbf{0.019}$\pm$0.001 \\ 
5  & 0.393$\pm$0.001 & 0.044$\pm$0.001 & 0.025$\pm$0.002 & \textbf{0.018}$\pm$0.001 \\ 
6  & 0.293$\pm$0.001 & 0.049$\pm$0.002 & 0.035$\pm$0.002 & \textbf{0.027}$\pm$0.001 \\ 
7  & 0.275$\pm$0.002 & 0.052$\pm$0.001 & 0.033$\pm$0.002 & \textbf{0.019}$\pm$0.001 \\ 
8  & 0.182$\pm$0.001 & 0.052$\pm$0.001 & 0.040$\pm$0.001 & \textbf{0.021}$\pm$0.002 \\ 
9  & 0.294$\pm$0.001 & 0.060$\pm$0.007 & 0.043$\pm$0.004 & \textbf{0.039}$\pm$0.007 \\ 
10 & 0.448$\pm$0.001 & 0.064$\pm$0.009 & \textbf{0.045}$\pm$0.007 & 0.047$\pm$0.006 \\
11 & 0.401$\pm$0.013 & 0.246$\pm$0.032 & \textbf{0.169}$\pm$0.011 & 0.210$\pm$0.026  \\ 
12 & 0.448$\pm$0.022 & 0.247$\pm$0.032 & \textbf{0.171}$\pm$0.014 & 0.193$\pm$0.005 \\ 
\bottomrule
\end{tabular}
}
\end{table}

\paragraph{Results.} From the table, we can see that both versions of our IGNR outperform the baselines across all graphons. \textbf{SAS} has the worst errors because it first zero-pads the input graphs to enforce them to have the same size before estimating the graphon. The padding confuses the graph alignment step, which is an important step for \textbf{SAS} and other classical graphon learning methods. \textbf{GWB} performs better than \textbf{SAS} because due to the usage of GW distance it neither requires padding, nor requires the input graphs to be aligned. It uses the GW barycenter of the input graphs to estimate the graphon, but has to specify the resolution $K$ for the estimate. We follow the original paper, and set $K=\lfloor\frac{N_{\max}}{\log N_{\max}}\rfloor$, where $N_{\max}$ is the size of the largest graph among the input. One may suspect that increasing $K$ can lower the error for \textbf{GWB}, but we show in the supplement that increasing $K$ beyond the suggested value actually increases the error. Our IGNR allows the input graphs to be un-aligned, does not require padding, and does not have any resolution parameter. The resolution-free representation of graphon helps IGNR to achieve the best performance. Among the two versions of IGNR, \textbf{IGNR-pg} performs better than \textbf{IGNR-cg} for the easy-to-align cases (Graphons 0-8); whereas \textbf{IGNR-cg} shows advantage over \textbf{IGNR-pg} for the cases (Graphons 10-12) where the degree of the ground truth graphon ($d(u)=\int_0^1 W(u,v)dv$) does not admit any strictly increasing ordering.


\subsection{c-IGNR on learning parameterized graphons}
\label{subsec:graphonAE} 


\paragraph{Experiments setup.} Motivated by IGNR's superior performance on learning single graphons, we move to the more challenging task of learning a family of graphons parameterized by $\alpha$. We consider two synthetic datasets of graphs generated by parameterized graphons. In the first scenario ($S_1$), we consider graphons with the shape of a stochastic block model composed of two-block, and the governing parameter $\alpha$ determines the size ratio between the two blocks (see Figure 2(a) top). Formally, we let
\begin{equation}
\begin{split}
    W_{\alpha}(x,y)=0.8\mathds{1}_{[0,\alpha]^2}(x,y)+0.8\mathds{1}_{[1-\alpha,1]^2}(x,y)\\
+0.1\mathds{1}_{[0,1]^2}(x,y)
\end{split}
\end{equation}
and we choose $\alpha$ to be in the range $\alpha \in [0.1,0.5]$. 
We generate 600 graphs from $W_{\alpha}(x,y)$ under the deterministic setting, with $\alpha$ randomly sampled from $\Uniform([0.1,0.5])$, and graph sizes randomly sampled from $\{50,51,...,79\}$ with equal probability. In the second scenario ($S_2$), we consider graphons that generate noisy ring graphs, and the parameter $\alpha$ determines the thickness (``noisiness") of the ring (see Figure 2(b) top). Formally, we let 
\begin{equation}
\begin{split}
    W_{\alpha}(x,y)=0.9\exp{((-y^2-(x-1)^2)/\alpha^2)}+\\
    0.9\exp{((-(y-1)^2-x^2)/\alpha^2)}+\\
    0.9\exp{(-((\sin\left(\frac{3}{4}\pi\right)x+\cos\left(\frac{3}{4}\pi\right)y)/\alpha)^2)}
\end{split}    
\end{equation}
and we choose $\alpha$ to be in the range $\alpha \in [0.05,0.15]$. 100 graphs are generated from $W_{\alpha}(x,y)$ under the deterministic setting, with $\alpha$ randomly sampled from $\Uniform([0.05,0.15])$, and graph sizes randomly sampled from $\{50,51,...,59\}$ with equal probability. In both settings, we assume no self-loops and generate undirected graph from the graphon by only sampling the upper-triangular part of the graphon. We use the adjacency matrix of the graph to represent its structure and consider a uniform measure/histogram on the nodes.

\begin{figure}[h]
\includegraphics[scale=0.38]{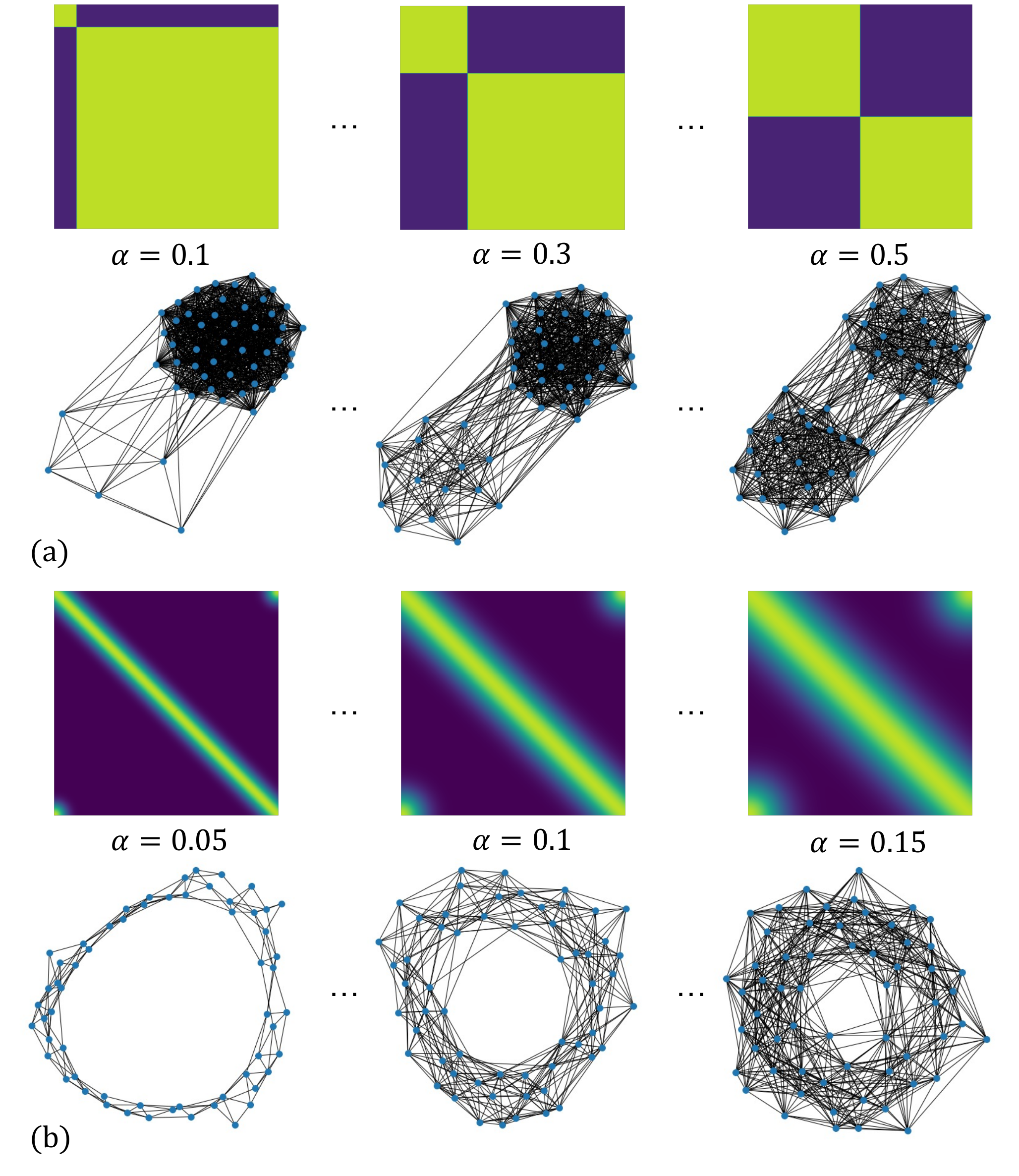}
\caption{Illustration of synthetic parameterized graphons (top) and corresponding graphs (bottom) generated from the graphons for $S_1$ in (a), and $S_2$ in (b).}
\end{figure}

\paragraph{Qualitative result.}
We first train our c-IGNR (within the AE framework) using data from $S_1$. To visually verify that c-IGNR is able to capture the ratio parameter $\alpha$, we train our model with latent dimension $d=2$, and visualize the latent codes in 2D colored by the value of $\alpha$ in Figure 3(a). We can see that the latent codes nicely capture the changing $\alpha$ in a 1D manifold. 

Interestingly, once the AE is trained, we can treat the learned decoder itself, i.e. our c-IGNR, as a separate model that can generate graphs of arbitrary sizes, including sizes not included in the training data, and of similar structure to the training graphs. This is because by feeding c-IGNR with a latent code, we obtain a graphon represented by IGNR, and can thus generate arbitrary sized graphs from the IGNR. We illustrate this in Figure 3(a), where we select a latent code (the gray star) in the low (high) $\alpha$ region from the latent space, and use this latent code to generate two graphs of sizes not observed in the training data. The newly generated graphs also exhibit a two-blocks structure, with the ratio between block sizes similar to that of the selected latent code. We repeat the same analysis with data from $S_2$ and present the result in Figure 3(b). The latent codes again capture the changing parameter $\alpha$, and new graphs of novel sizes can be generated from the c-IGNR.


\begin{figure}[h]
\includegraphics[scale=0.33]{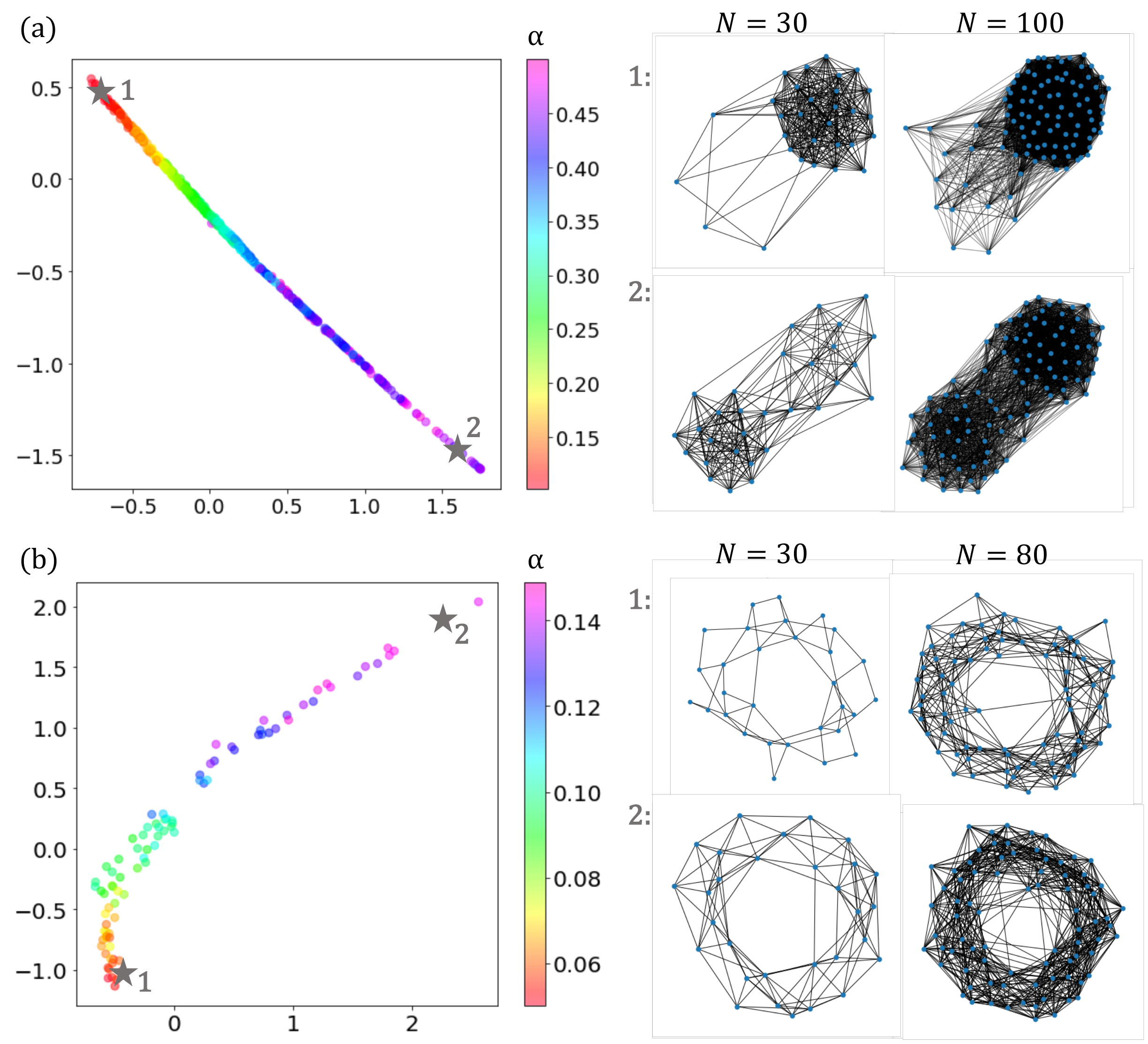}
\caption{Left: 2-dimensional latent code learned by c-IGNR. Right: Using the latent codes represented by the gray stars in the left embedding plot, we can generate new graphs of sizes not included in the training data from c-IGNR. (a)$S_1$, (b)$S_2$.}
\end{figure}

\paragraph{Quantitative result.}
To quantitatively evaluate the performance of our model on learning parameterized graphons, we split the data into a training set and a hold-out testing set. After the model is trained on the training set, we estimate the graphon for each graph in the testing set, and report the mean squared GW distance between each estimated graphon and the ground-truth graphon (at resolution $1000\times 1000$) that generated the corresponding testing graph. 

To demonstrate the merit of c-IGNR as an efficient (in terms of the number of parameters) and resolution-free decoder, we compare it to a discrete baseline which uses the same encoder, but the decoder is replaced by an MLP that maps the latent code to a fixed $K \times K$ matrix with values in $[0,1]$. In other words, the discrete baseline's output matrix at resolution $K$ is treated as the piece-wise constant approximation to the ground-truth graphon, and $K$ is a resolution parameter. In general, the performance of the discrete baseline improves with larger $K$, but its number of parameters grows quickly with $K$ ($O(K^2)$); the discrete baseline's error is still larger than the error of c-IGNR, when its number of parameters far exceeds that of c-IGNR (see supplemental material). 

We also compare our method with state-of-the-art GW method, Graph Dictionary Learning (GDL)~\cite{vincent2021online}. We train GDL on the same training set to obtain the dictionary atoms, and use the learned dictionary atoms to reconstruct each testing set graph. The reconstructed graphs, which are weighted adjacency matrices taking value in $[0,1]$, are treated as piece-wise approximation to the ground-truth graphon. We sweep the number of atoms and size of the atoms (see supplemental material) and report the lowest test error for GDL across parameter configurations. We run each experiment five times and report the mean and standard deviation of error on the test set. The result is summarized in Table 2. Our c-IGNR outperforms both the discrete baseline and GDL.

\begin{table}[ht]
\caption{Errors of learning parameterized graphons}
    \centering\small
    \begin{tabular}{llll}
        \toprule
         & c-IGNR                           & Discrete                         & GDL    \\ 
         \midrule
$S_1$ & \textbf{0.022}$\pm$0.001 & 0.038$\pm$0.001 & 0.052$\pm$0.002  \\ 
$S_2$ & \textbf{0.024}$\pm$0.004 & 0.037$\pm$0.005 & 0.029$\pm$0.003  \\ 
        \bottomrule
    \end{tabular}
\end{table}

\subsection{c-IGNR on real data for classification}
\label{subsec:classification}

Finally, we show that on real datasets, the latent codes learned by our c-IGNR (within the AE framework) can also be useful for a graph classification task. To this end, we consider the two well-known benchmark datasets IMDB-B and IMDB-M \cite{yanardag2015deep}, which contain social network graphs without node attributes. We compare our methods with two GW baselines, GDL~\cite{vincent2021online} and GWF~\cite{xu2020gromov}. For GDL, the weights that correspond to the linear factorization of the graph w.r.t the dictionaries give the graph embedding vector that can be used for classification. For GWF, the weights that correspond to the barycenter representation of the graph serve as the embedding vector for classification. Both our methods and the GW baselines are first trained in an unsupervised manner to obtain the graph embedding/latent vectors, and then an SVM classifier on those vectors are trained. We report the 10-fold cross validation classification accuracy. The $C$ parameter for the SVM are cross validated within $C\in\{10^{-4},...,10^4\}$. The result is summarized in Table 3. We can see that our c-IGNR can achieve meaningful latent representations for graph classification.

\begin{table}[h]
\caption{Classification accuracy (\%)}
    \centering\small
    \begin{tabular}{llll}
        \toprule
         & c-IGNR                           & GWF                         & GDL    \\ 
         \midrule
IMDB-B & \textbf{72.00}$\pm$0.42 & 61.00$\pm$2.43 &  70.11$\pm$3.13 \\ 
IMDB-M & 48.48$\pm$0.56 & 40.35$\pm$0.79 &   \textbf{49.01}$\pm$3.66\\ 
        \bottomrule
    \end{tabular}
\end{table}

\section{CONCLUSION}
In this paper, we proposed IGNR, a novel framework to model and learn graphons leveraging implicit neural representations and the Gromov-Wasserstein distance. We demonstrated that our IGNR excels in both classical and more complex settings of graphon learning. 
It would be interesting to provide precise theoretical statements regarding how implicit neural representations can approximate graphons (or families of well-behaved graphons), which we aim to do in the future.

One limitation of the present work is that graphs sampled from graphons are usually dense. It will be interesting to explore how to allow the generation of sparser graphs. One possibility is to output a certain ``sparsification'' factor together with the generated graphon. 

\add{
\section{ACKNOWLEDGEMENTS}
This work was supported by NSF under CCF-2217033 and CCF-2112665.
}

\bibliography{refs}

\appendix
\onecolumn

\section{Implementation Details of IGNR and c-IGNR}

\add{Our code is available at: https://github.com/Mishne-Lab/IGNR}

We detail the parameter setting for IGNR and c-IGNR in this section. The IGNR for learning a single graphon is an MLP with 3 hidden layers. Each hidden layer contains 20 hidden units. Sine activation is used for each hidden layer, and the output layer uses sigmoid activation. The IGNR takes 2-dimensional coordinate input and outputs 1-dimensional edge probability (also see Figure 1 (c) left in the main text). For the c-IGNR, its synthesis network is an MLP with 3 hidden layers, and each hidden layers contain 48, 36, 24 hidden units respectively. Sine activation is used for each hidden layer, and the output layer uses sigmoid activation. The synthesis network of c-IGNR takes 2-dimensional coordinate input and outputs 1-dimensional edge probability. The modulation network of c-IGNR is also an MLP with 3 hidden layers, and each hidden layers contain 48, 36, 24 hidden units respectively. ReLU activation is used for each hidden layer. The modulation network takes the latent code (dimension depending on experiments, tested in $\{2,3,16,32,64\}$) as input, and does not have further output layer (also see Figure 1 (c) right in the main text). To train the c-IGNR within an auto-encoder framework, we have to specify the encoder module. We use a 3-layer GIN \cite{xu2018powerful} as our encoder, and apply global average pooling to obtain the graph embedding that is used as the latent code for the input graph. 

We implement our model in Python using the POT library (Python Optimal Transport) \cite{flamary2021pot} and Pytorch library \cite{paszke2019pytorch}. We optimize the parameters of our model using Adam Optimizer. Computation of the OT coupling matrices ($T_i$) are based on POT. Note that for the single graphon learning task, because the number of input graphs is small, we test two versions of our IGNR that use the conditional gradient (CG) algorithm and the proximal gradient (PG) algorithm to solve for the OT matrices, respectively. For c-IGNR, we compute OT matrices only using the CG algorithm, because a) in the more general and realistic settings, it is unlikely that the input graphs can admit strictly increasing node degree ordering (which is where PG showing slight advantage over CG); and b) when the input number of graphs is much larger than that of the simple setting, it becomes more computationally expensive to use PG, because it has to keep a current estimation of the OT coupling matrix for each training graph at each iteration. 

\add{
The current computational bottleneck for (c-)IGNR is the GW loss computation with the POT package
 (complexity is up to $O(N^2M)$ where $N$ is the largest graph size in the dataset and $M\leq N$ is the graph reconstruction size). Learning IGNR (Sec 4.1) on a single core of a CPU(Intel Xeon Gold 6230@2.10GHz) takes an average of 16.8s; learning c-IGNR on IMDB-B on a single Nvidia Quadro RTX 8000 GPU takes an average of 121.9s.
 Our future work aims to improve this.
}

\section{Additional Experimental Information for learning single graphons}

\subsection{Functions used for the single graphon learning task}
Table 4 below shows the definitions of the 13 ground truth graphons used in the single graphon learning task (see section 4.1 of the main text).
\setlength\extrarowheight{4pt}
\begin{table}[ht]
\caption{Ground truth graphons corresponding to each index in Table 1 of the main text.}
    \centering\small
    \begin{tabular}{llll}
        \toprule
         & $W(x,y)$   \\ 
         \midrule
0 &  $xy$  \\ 
1 &  $\exp(-(x^{0.7}+y^{0.7}))$  \\ 
2 &  $\frac{1}{4}(x^2+y^2+\sqrt{x}+\sqrt{y})$\\
3 &  $\frac{1}{2}(x+y)$\\
4 &  $(1+\exp(-2(x^2+y^2)))^{-1}$\\
5 &  $(1+\exp(-\max\{x,y\}^2-\min\{x,y\}^4))^{-1}$\\
6 &  $\exp(-\max\{x,y\}^{0.75})$\\
7 &  $\exp(-\frac{1}{2}(\min\{x,y\}+\sqrt{x}+\sqrt{y}))$\\
8 &  $\log(1+\max\{x,y\})$\\
9 &  $|x-y|$\\
10 & $1-|x-y|$\\
11 & $0.8\mathbf{I}_2\otimes\mathds{1}_{[0,\frac{1}{2}]^2}$\\
12 & $0.8(1-\mathbf{I}_2)\otimes\mathds{1}_{[0,\frac{1}{2}]^2}$\\
        \bottomrule
    \end{tabular}
\end{table}

\add{
\subsection{Effect of IGNR architecture}

In Figure 4, we show the result for one trial of the single graphon learning experiment where we compare IGNR with different numbers of hidden units (labeled by the legend) with the strongest baseline \textbf{GWB}. We can see that the performance of IGNR is not very sensitive to the choice of specific configurations, in the sense that all different configurations outperform \textbf{GWB}. Note that in the extreme case of learning the graphon from only one graph input, IGNR has the risk of ``overfitting" to the particular entries of the input graph adjacency matrix, especially when the network size is large (smaller and simpler networks would impose implicit regularization and mitigate the "overfitting" effect). However, since in our experiments we are learning the graphon from multiple graphs, IGNR would learn the average edge probability among the graphs and thus avoid the problem of ``overfitting" to a specific input graph. Consequently, we observe in Figure 4 that IGNR performs well across networks sizes. 
}
\begin{figure}[h]
\includegraphics[scale=0.6]{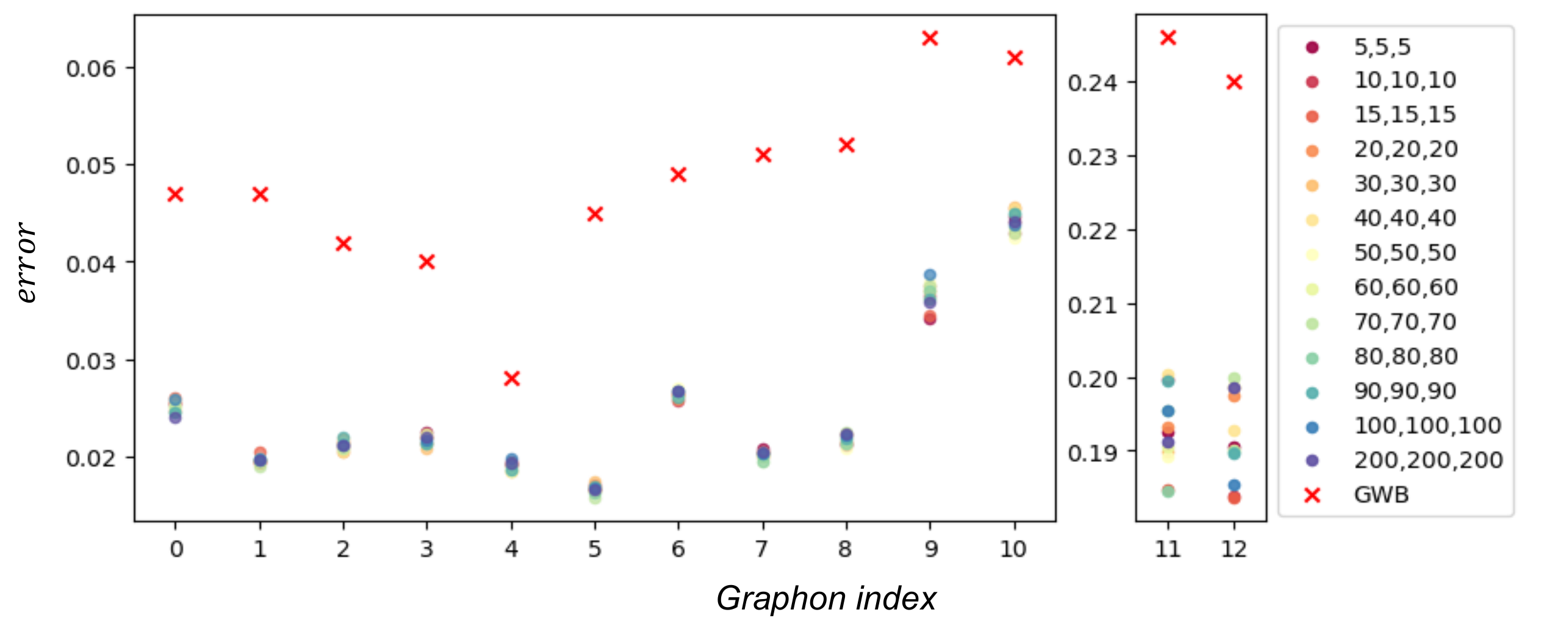}
\caption{Effects of IGNR architectures (labelled by the legend on the right) on the single graphon learning task. x-axis corresponds to the indices of different graphons, and y-axis indicates the error.}
\end{figure}

\add{
\subsection{Additional evaluation metric}
In Table 1, we used the 2-order GW distance between the ground-truth graphon and the estimated graphon as an evaluation metric because the GW distance is permutation invariant and does not require the two graphons under comparison to be aligned. Graphons 0-8, however, can be easily identified/aligned by sorting the node degrees (their degree function can be ordered in strictly increasing order). Therefore, we can use an additional metric, the mean squared error (MSE), to evaluate the graphon reconstruction for these easy to align graphons (evaluate the MSE after sorting the degrees of the two graphons under comparison). Table 5 summarize the MSE for Graphons 0-8 to accompany the results in Table 1. We can see that IGNR outperforms the strongest baseline under this additional metric. 
}
\setlength\extrarowheight{2pt}
\begin{table}[ht]
\caption{MSE of learning single graphons (0-8)} 
\centering \tiny
\resizebox{8cm}{!}{%
\begin{tabular}{lllll} 
\toprule
\addlinespace[8pt]
 &
  \cellcolor[HTML]{FFFFFF}\textbf{IGNR-pg} &
  \cellcolor[HTML]{FFFFFF}\textbf{IGNR-cg} &
  \cellcolor[HTML]{FFFFFF}\textbf{GWB} \\ 
\midrule
0 & 25.410±3.544  & 31.800±4.478   & 47.137±6.344 \\ 
1 & 19.543±2.475 & 34.639±2.254 & 47.150±1.390  \\ 
2 & 21.712±2.577 & 27.039±2.969 & 42.294±3.686 \\ 
3 & 21.537±3.239 & 26.116±3.515 & 39.866±4.458 \\ 
4 & 18.906±1.586 & 21.598±2.935 & 27.764±2.102 \\ 
5 & 17.222±1.371 & 24.280±2.349  & 45.058±2.019 \\ 
6 & 26.656±1.371 & 34.484±2.292 & 50.064±1.374 \\ 
7 & 19.390±1.409  & 33.868±1.449 & 51.519±1.257 \\ 
8 & 21.823±1.869 & 41.281±2.022 & 52.609±1.576 \\ 
\bottomrule
\end{tabular}
}
\end{table}

\subsection{Comparison with additional classical graphon learning methods}
In Table 6, we extend the results in Table 1 of the main text and show the errors of single graphon learning with additional classical baselines --- stochastic block approximation (\textbf{SBA}) \cite{airoldi2013stochastic}, largest gap (\textbf{LG}) \cite{channarond2012classification}, matrix completion (\textbf{MC}) \cite{keshavan2010matrix}, and universal singular value thresholding (\textbf{USVT}) \cite{chatterjee2015matrix}. We can see that similar to \textbf{SAS}, those classical methods suffer from aligning and zero-padding, and cannot outperform our method and \textbf{GWB}.

\setlength\extrarowheight{2pt}
\begin{table}[ht]
\caption{Errors of learning single graphons (each index corresponds to a different graphon)} 
\centering \tiny
\resizebox{\columnwidth}{!}{%
\begin{tabular}{lllllllll} 
\toprule
\addlinespace[8pt]
 &
  \textbf{SBA}&
  \textbf{LG}&
  \textbf{MC}&
  \textbf{USVT}&
  \textbf{SAS} &
  \cellcolor[HTML]{FFFFFF}\textbf{GWB} &
  \cellcolor[HTML]{FFFFFF}\textbf{IGNR-cg} &
  \cellcolor[HTML]{FFFFFF}\textbf{IGNR-pg} \\ 
\midrule
0  & 0.265$\pm$0.002 & 0.178$\pm$0.004 & 0.108$\pm$0.004 & 0.18$\pm$0.004  & 0.173$\pm$0.005 & 0.049$\pm$0.008 & 0.034$\pm$0.005 & \textbf{0.026}$\pm$0.003 \\ 
1  & 0.235$\pm$0.002 & 0.200$\pm$0.002 & 0.204$\pm$0.002 & 0.205$\pm$0.002 & 0.197$\pm$0.002 & 0.047$\pm$0.001 & 0.035$\pm$0.002 & \textbf{0.020}$\pm$0.002  \\
2  & 0.348$\pm$0.003 & 0.307$\pm$0.003 & 0.310$\pm$0.003 & 0.311$\pm$0.003 & 0.303$\pm$0.003 & 0.044$\pm$0.004 & 0.028$\pm$0.003 & \textbf{0.023}$\pm$0.003 \\ 
3  & 0.359$\pm$0.003 & 0.309$\pm$0.003 & 0.312$\pm$0.003 & 0.312$\pm$0.003 & 0.305$\pm$0.003 & 0.042$\pm$0.004 & 0.028$\pm$0.004 & \textbf{0.023}$\pm$0.004 \\ 
4  & 0.468$\pm$0.002 & 0.473$\pm$0.002 & 0.475$\pm$0.002 & 0.475$\pm$0.002 & 0.468$\pm$0.002 & 0.029$\pm$0.002 & 0.023$\pm$0.003 & \textbf{0.019}$\pm$0.001 \\ 
5  & 0.383$\pm$0.001 & 0.397$\pm$0.001 & 0.400$\pm$0.001 & 0.401$\pm$0.001 & 0.393$\pm$0.001 & 0.044$\pm$0.001 & 0.025$\pm$0.002 & \textbf{0.018}$\pm$0.001 \\
6  & 0.312$\pm$0.002 & 0.296$\pm$0.001 & 0.300$\pm$0.001 & 0.301$\pm$0.001 & 0.293$\pm$0.001 & 0.049$\pm$0.002 & 0.035$\pm$0.002 & \textbf{0.027}$\pm$0.001 \\ 
7  & 0.294$\pm$0.002 & 0.278$\pm$0.002 & 0.282$\pm$0.001 & 0.283$\pm$0.001 & 0.275$\pm$0.002 & 0.052$\pm$0.001 & 0.033$\pm$0.002 & \textbf{0.019}$\pm$0.001 \\
8  & 0.191$\pm$0.001 & 0.184$\pm$0.001 & 0.189$\pm$0.001 & 0.189$\pm$0.001 & 0.182$\pm$0.001 & 0.052$\pm$0.001 & 0.040$\pm$0.001 & \textbf{0.021}$\pm$0.002 \\
9  & 0.308$\pm$0.001 & 0.296$\pm$0.001 & 0.293$\pm$0.001 & 0.294$\pm$0.001 & 0.294$\pm$0.001 & 0.060$\pm$0.007 & 0.043$\pm$0.004 & \textbf{0.039}$\pm$0.007 \\
10 & 0.458$\pm$0.001 & 0.452$\pm$0.001 & 0.450$\pm$0.001 & 0.451$\pm$0.001 & 0.448$\pm$0.001 & 0.064$\pm$0.009 & \textbf{0.045}$\pm$0.007 & 0.047$\pm$0.006 \\
11 & 0.464$\pm$0.001 & 0.403$\pm$0.013 & 0.403$\pm$0.012 & 0.403$\pm$0.012 & 0.401$\pm$0.013 & 0.246$\pm$0.032 & \textbf{0.169}$\pm$0.011 & 0.210$\pm$0.026  \\
12 & 0.465$\pm$0.001 & 0.443$\pm$0.015 & 0.437$\pm$0.008 & 0.445$\pm$0.02  & 0.448$\pm$0.022 & 0.247$\pm$0.032 & \textbf{0.171}$\pm$0.014 & 0.193$\pm$0.005 \\
\bottomrule
\end{tabular}
}
\end{table}

\subsection{GWB and resolution parameter}
For \textbf{GWB}, the GW barycenter of the input graphs is used as the piece-wise constant approximation of the ground-truth graphon. According to \cite{xu2021learning}, the resolution of the GW barycenter is set as $K=\lfloor\frac{N_{\max}}{\log N_{\max}}\rfloor$, where $N_{\max}$ is the size of the largest graph among the input graphs. One may suspect that one can always increase the resolution $K$ to achieve better performance of \textbf{GWB}. In Figure 5 below, we can see that increasing the resolution $K$ beyond the recommended value ($K=36$) does not improve \textbf{GWB}'s performance. In contrast, our IGNR does not rely on any resolution parameter.

\begin{figure}[h]
\includegraphics[scale=0.38]{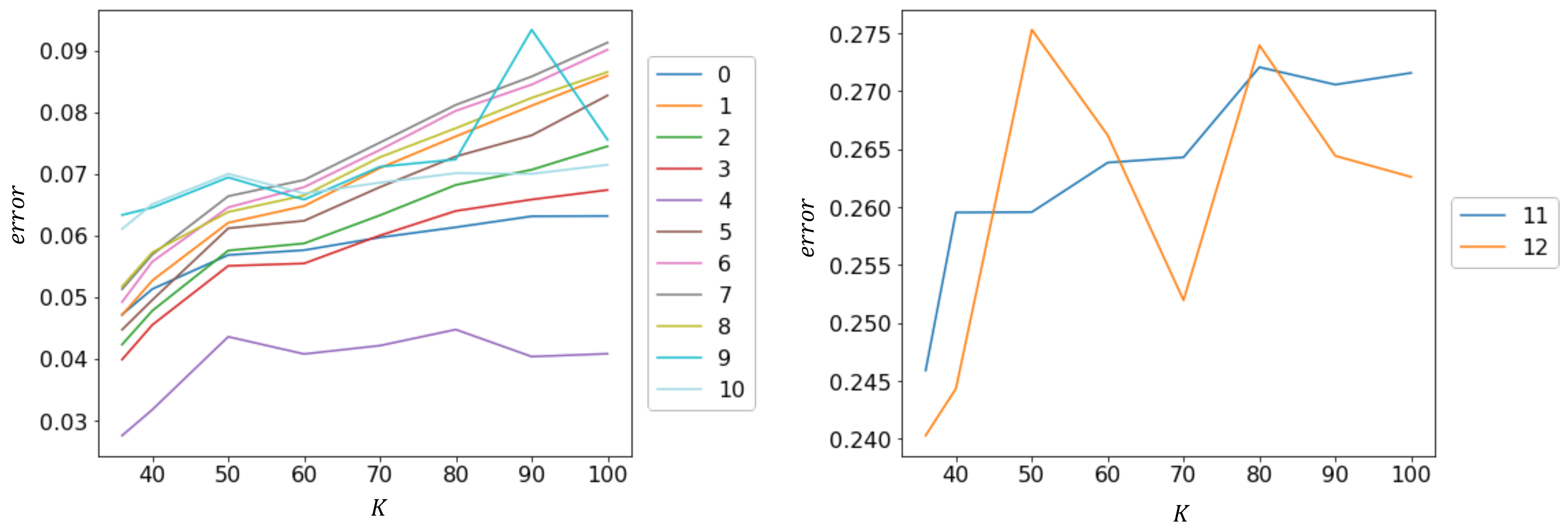}
\caption{Effects of the resolution parameter $K$ for \textbf{GWB}. x-axis indicates resolution, and y-axis indicates the mean error. Indices in the legend correspond to the graphons in Table 1.}
\end{figure}

\section{Additional Experimental Information for learning parameterized family of graphons}

\subsection{Parameter comparison with discrete baseline}
To demonstrate that within an auto-encoder framework, c-IGNR is not only an effective graphon decoder, but is also efficient in terms of its parameter usage, we construct a discrete baseline decoder (D). Concretely, the decoder D is an MLP that takes the latent graph embedding as the input and outputs a fixed $K \times K$ symmetric matrix with values in $[0,1]$, which is treated as the piece-wise constant approximation of the ground truth graphon at resolution $K$. The same encoder module is used for both c-IGNR and the discrete baseline D. Table 7 and 8 below demonstrate the number of parameters in D and c-IGNR under different configurations and their respective test set error for learning parameterized graphons under $\mathcal{S}_1$ and $\mathcal{S}_2$. The format for displaying configuration is [model]-[latent code dimension]-[dimension of hidden layers in MLP]. We can see that as a general trend, the error for D decreases as the resolution parameter $K$ increases. However, because the configure of D depends on $K$, the number of parameters in D increases quickly ($O(K^2)$) and far exceed the number of parameters in c-IGNR when its error is still much larger than that of c-IGNR. In contrast, the number of parameters in c-IGNR
does not depend on the output resolution due to its resolution-free representation.

\begin{table}[ht]
\parbox{0.01\linewidth}{
\centering
\resizebox{235pt}{!}{
\begin{tabular}{lllll}
    \toprule
    \multirow{2}{*}{Models} & \multicolumn{4}{c}{resolution $K$}\\
    \cmidrule{2-5} \\
    {} & 18 & 24 & 36 & 48 \\
    \midrule
D-16-{[}32,64,128{]} & 30336 & 46080 & 91392 & 155136 \\ 
D-16-{[}32,64{]}     & 12352 & 20224 & 42880 & 74752  \\ 
D-16-{[}16,16{]}     & 2960  & 4928  & 10592 & 18560  \\ 
D-2-{[}32,64,128{]}  & 29888 & 45632 & 90944 & 154688 \\ 
D-2-{[}32,64{]}      & 11904 & 11976 & 42432 & 74304  \\ 
D-2-{[}16,16{]}      & 2736  & 4047  & 10368 & 18336  \\ 
    \midrule
    c-IGNR-16-[48,36.24] & \multicolumn{4}{c}{7032} \\
    c-IGNR-2-[48,36,24] & \multicolumn{4}{c}{5520} \\
    \bottomrule
\end{tabular}}
\caption{Comparing number of parameters (left) and mean errors (right) for learning parameterized graphons between c-IGNR and the discrete baseline decoder (D) for dataset in $\mathcal{S}_1$.}
}
\hfill
\parbox{0.45\linewidth}{
\centering
\resizebox{222pt}{!}{
\begin{tabular}{lllll}
    \toprule
    \multirow{2}{*}{Models} & \multicolumn{4}{c}{resolution $K$}\\
    \cmidrule{2-5} \\
    {} & 18 & 24 & 36 & 48 \\
    \midrule
D-16-{[}32,64,128{]} & 0.067 & 0.055 & 0.044 & 0.040 \\ 
D-16-{[}32,64{]}     & 0.072 & 0.056 & 0.044 & 0.038 \\ 
D-16-{[}16,16{]}     & 0.067 & 0.056 & 0.045 & 0.039 \\ 
D-2-{[}32,64,128{]}  & 0.081 & 0.059 & 0.052 & 0.069 \\
D-2-{[}32,64{]}      & 0.071 & 0.062 & 0.048 & 0.058 \\ 
D-2-{[}16,16{]}      & 0.077 & 0.062 & 0.047 & 0.051 \\ 
    \midrule
    c-IGNR-16-[48,36.24] & \multicolumn{4}{c}{0.022} \\
    c-IGNR-2-[48,36,24] & \multicolumn{4}{c}{0.025} \\
    \bottomrule
\end{tabular}
}}
\end{table}

\begin{table}[ht]
\parbox{0.01\linewidth}{
\centering
\begin{tabular}{llll}
    \toprule
    \multirow{2}{*}{Models} & \multicolumn{3}{c}{resolution $K$}\\
    \cmidrule{2-4} \\
    {} & 40 & 50 & 60 \\
    \midrule
    D-32-{[}32,64{]} & 52992 & 81472 & 116352 \\
    D-32-{[}32,32{]} & 27008 & 41248 & 58688  \\
    D-2-{[}32,64{]}  & 52032 & 80512 & 115392 \\
    D-2-{[}32,32{]}  & 26048 & 40288 & 57728 \\
    \midrule
    c-IGNR-32-[48,36.24] & \multicolumn{2}{c}{8760} \\
    c-IGNR-2-[48,36,24] & \multicolumn{2}{c}{5520} \\
    \bottomrule
\end{tabular}
\caption{Comparing number of parameters (left) and mean errors (right) for learning parameterized graphons between c-IGNR and the discrete baseline decoder (D) for dataset in $\mathcal{S}_2$.}
}
\hfill
\parbox{0.45\linewidth}{
\centering
\begin{tabular}{llll}
    \toprule
    \multirow{2}{*}{Models} & \multicolumn{3}{c}{resolution $K$}\\
    \cmidrule{2-4} \\
    {} & 40 & 50 & 60 \\
    \midrule
D-32-{[}32,64{]} & 0.073 & 0.042 & 0.037 \\
D-32-{[}32,32{]} & 0.072 & 0.042 & 0.038 \\
D-2-{[}32,64{]}  & 0.073 & 0.039 & 0.040 \\
D-2-{[}32,32{]}  & 0.073 & 0.046 & 0.040 \\
    \midrule
    c-IGNR-32-[48,36.24] & \multicolumn{2}{c}{0.024} \\
    c-IGNR-2-[48,36,24] & \multicolumn{2}{c}{0.026} \\
    \bottomrule
\end{tabular}
}
\end{table}

\subsection{Comparison with GDL}
The \textbf{GWB} method for learning single graphons is not applicable for learning parameterized family of graphons. Therefore to compare with existing GW methods under this more general setting (of learning parameterized graphons), we choose to compare with \textbf{GDL} \cite{vincent2021online}. In particular, we learn the dictionary graphs at resolution $K$ from the training set, and then use this learned set of dictionaries to reconstruct each graph in the testing set. The reconstructed graphs, represented as $K \times K$ weighted adjacency matrices taking values in $[0,1]$ are treated as the piecewise constant approximations of the ground-truth graphons at resolution $K$. Note that compared with our c-IGNR, GDL is not ``inductive"---when computing reconstructed graphs (and corresponding weights for each dictionary) in the testing set, one has to independently solve an unmixing problem for each graph in the testing set, which significantly increases evaluation time, especially when the testing set is large. Our c-IGNR is ``inductive" in the sense that, once the model is trained from the training set, we can directly apply it to new graph data (without solving any new optimization problems), and obtain new graph embeddings and graphon reconstructions (at arbitrary resolutions). For \textbf{GDL}, we use its implementation in the POT python library. The number of dictionary atoms is tested among $\{2,16,32\}$, and the dictionary size is tested among $\{10,20,30,40,50,60,70,80,90,100\}$. The best error achieved for \textbf{GDL} across parameters is reported in the main text.

\vfill

\end{document}